\documentclass[]{gtech}
\usepackage{amssymb}
\usepackage{multirow}
\usepackage{bigdelim}
\usepackage{todonotes}
\usepackage{longtable}
\usepackage{tabularray}
\usepackage{wrapfig}
\usepackage[most]{tcolorbox} 
\usepackage{xcolor}
\usepackage{url}

\renewcommand{\title}[1]{\newcommand{\titlelist}{{\huge\fontfamily{optimistic}\selectfont #1}}}

\newcommand{\model}{\texttt{Ming-Lite-Uni}}
\newcommand{\modellite}{\texttt{Ming-Lite-Uni}}

\definecolor{prompt}{HTML}{5f84e4}
\definecolor{img}{HTML}{820100}

\usepackage{arydshln}

\definecolor{CQColor}{rgb}{0.0,0.0,1.0} 

\definecolor{TSColor}{rgb}{0.5,0.0,0.8} 

\definecolor{CQRColor}{rgb}{1.0,0.0,1.0} 

\usepackage{colortbl}
\usepackage{amssymb}
\usepackage{pifont}
\usepackage{booktabs,multirow}
\usepackage{makecell}
\usepackage{tabulary}
\usepackage{fontawesome5}
\usepackage{bbding}
\usepackage{multicol}

\newlength\savewidth\newcommand\shline{\noalign{\global\savewidth\arrayrulewidth
  \global\arrayrulewidth 1pt}\hline\noalign{\global\arrayrulewidth\savewidth}}

\title{\textcolor[HTML]{0369ff}{Ming}-Lite-Uni: Advancements in Uni{f}{i}ed Architecture for Natural Multimodal Interaction}

\author[*]{Inclusion AI, Ant Group}

\contribution[*]{See Contributions section (Sec.~\ref{sec:contri}) for full author list.}

\abstract{\fontsize{11pt}{12pt} \textit{We introduce \model{}, an open-source multimodal framework featuring a newly designed unified visual generator and a native multimodal autoregressive model tailored for unifying vision and language.
Specifically, this project provides an open-source implementation of the integrated MetaQueries and M2-omni framework, while introducing the novel \textbf{multi-scale learnable tokens} and \textbf{multi-scale
representation alignment strategy}. By leveraging a fixed MLLM and a learnable diffusion model, \model{} enables native multimodal AR models to perform both text-to-image generation and instruction based image editing tasks, expanding their capabilities beyond pure visual understanding.
Our experimental results demonstrate the strong performance of \model{} and illustrate the impressive fluid nature of its interactive process. 
All code and model weights are open-sourced to foster further exploration within the community. \textbf{Notably}, this work aligns with concurrent multimodal AI milestones - such as ChatGPT-4o with native image generation updated in March 25, 2025 - underscoring the broader significance of unified models like \model{} on the path toward AGI. 
\model{} is in alpha stage and will soon be further refined.
}}

\date{May 05, 2025\vspace{-1mm}}
\gtechdata[Code]{\url{https://github.com/inclusionAI/Ming/tree/Ming-Lite-Omni-Preview/Ming-unify}}

\begin{document}
\maketitle

\section{Introduction}
\label{sec:intro}

\begin{figure*}[t]
\centering
\includegraphics[width=\textwidth]{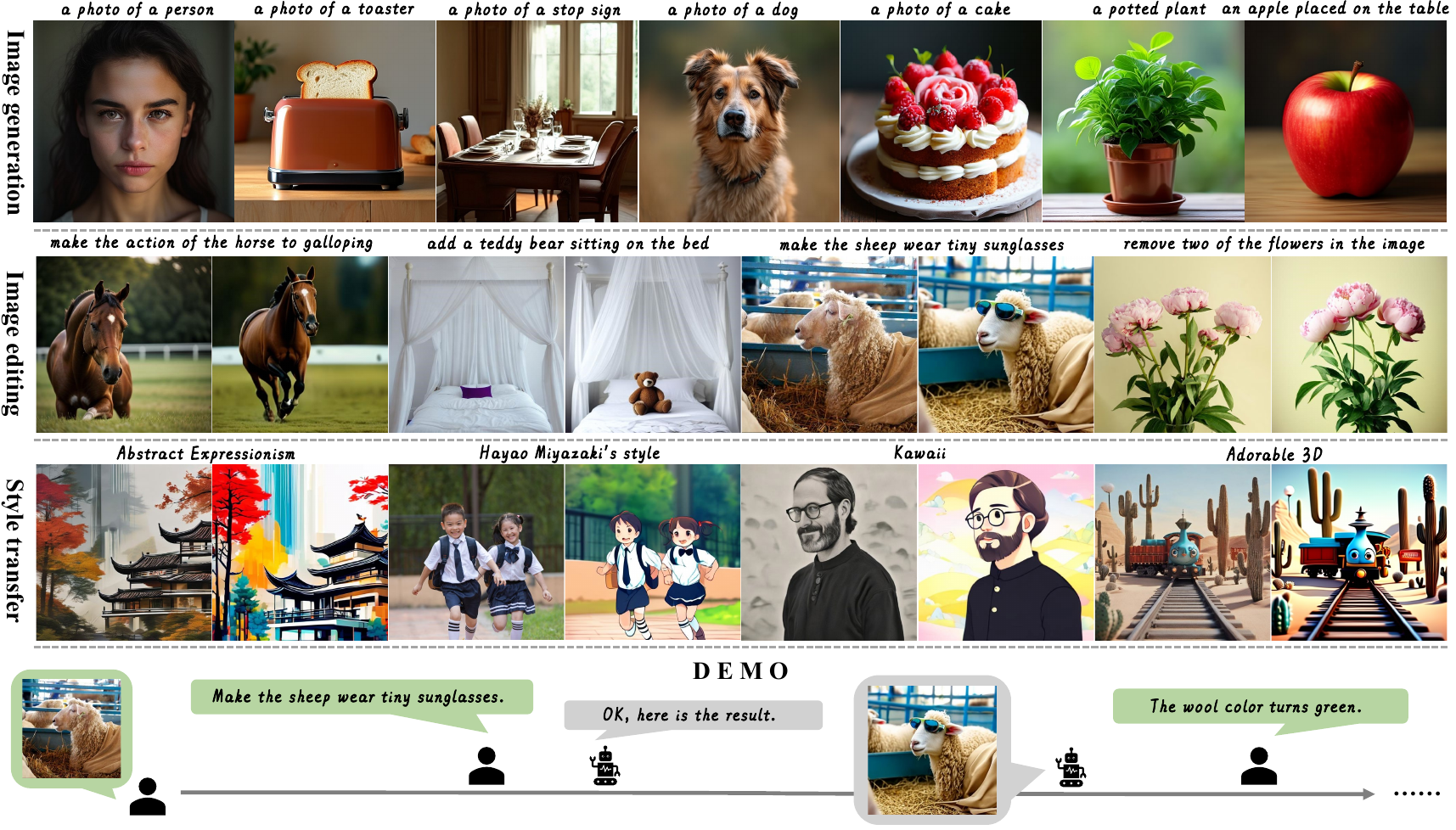}
\caption{\textbf{The output results and multimodal interactive demos of \model{}}. Our model supports basic multimodal chatting, text-to-image generation, image editing, and image style transfer based on textual instructions.}
\vspace{-2mm}
\label{fig:intro_image}
\end{figure*}

The release of GPT-4o~\cite{chatgpt4o} in March 2025, which introduced native image generation, has made the impact of unified models~\cite{tong2024metamorph,team2024chameleon,pan2025transfer} increasingly apparent. Users can now perform complex visual tasks such as image editing~\cite{gong2024check,feng2024ranni}, multi-view synthesis~\cite{shi2023zero123++}, style transfer~\cite{li2024styletokenizer}, and even 3D rendering~\cite{mildenhall2021nerf} purely through natural language conversations. 
These capabilities, once dependent on specialized models~\cite{gong2024uknow,wang2024framer,huang2024learning,tan2024animate}, can now be achieved fluently and with high quality, marking a major advance in perceived intelligence. 
In this context, our work pursues two primary objectives: (1) to demonstrate the strong potential of a unified auto-regressive multimodal model built upon the multi-scale learnable tokens with fine-tuned diffusion models, and (2) to accelerate community engagement by open-sourcing an alpha-version of our code and model weights.

A major challenge in unifying multimodal understanding and generation lies in the inconsistency of visual feature spaces. Recent models such as TokenFlow~\cite{qu2024tokenflow} and Janus~\cite{deepseek_janus} integrate diffusion-based decoders with token-based understanding models, achieving strong image generation but often at the cost of precise understanding. These methods prioritize pixel fidelity, leading to a mismatch between visual features and semantic understanding. 
\model{} adopts a lightweight bridging approach, serving as an open-source implementation and improvement over the integrated MetaQueries~\cite{pan2025transfer} and M2-omni~\cite{guo2025m2} framework. Unlike previous works, which focused on understanding performance and model structure, \model{} fixes the MLLM and fine-tunes the diffusion model through the newly designed multi-scale learnable tokens, multi-scale
representation alignment mechanism, and connector. 
In text-to-image generation and instruction based image editing tasks, autoregressive models excel at semantic understanding, providing robust contextual guidance. Meanwhile, finetuned diffusion models leverage multi-scale learnable tokens and tailored loss functions to achieve high-fidelity, fine-grained image synthesis.

A second core component of \model{} focuses on enhancing the visual generation capacity of its auto-regressive backbone, which is achieved by integrating a FlowMatching loss~\cite{esser2403scaling} directly into the separate diffusion model~\cite{tan2024mimir,shi2024motionstone,ma2024learning}.
Such an approach allows generation quality to improve in tandem with end-to-end training, enabling \model{} to effectively optimize the diffusion model while keeping the MLLM frozen. 
Additional efforts such as new multimodal AR architecture, scaling-up 
rule, pre-training and post-training strategy of the auto-regressive module help balance the model’s visual and language capacity, ensuring that \model{} remains a unified prototype with high potential and no compromises between modalities and tasks. Further improvements to the auto-regressive component are already underway and will be included in the next release.

Finally, we curate a multimodal dataset covering tasks such as image editing~\cite{tan2024ominicontrol} to train \model{}. Despite limited resources, \model{} exhibits strong control fluency and contextual understanding, handling diverse fine-grained tasks like image-to-text and text-to-image QA through natural language dialogue. All code and weights have been open-sourced, with a full experimental evaluation to follow in the next release.

\section{Approach}
\label{sec:method}

\model{} compresses image representations into a sequence of continuous tokens, which are combined with discrete text tokens and processed by a scaled auto-regressive Transformers~\cite{bai2025qwen25vltechnicalreport,lu2024deepseek} for end-to-end multimodal context learning. 
The generation capability is provided by an externally trainable diffusion model (\textit{e.g.,} SANA~\cite{xie2024sana}), conditioned on tokens produced by auto-regressive Transformers. 
In this Section, we first introduce our newly designed multi-scale learnable tokens and the multi-scale representation alignment strategy in Sec.~\ref{subsection:token}, followed by our previous designed Native Multimodal Auto-regressive Model in Sec.~\ref{subsection:ar}.

\subsection{\model{} with Multi-Scale Learnable Tokens}
\label{subsection:token}

\begin{figure*}[ht]
\centering
\includegraphics[width=\textwidth]{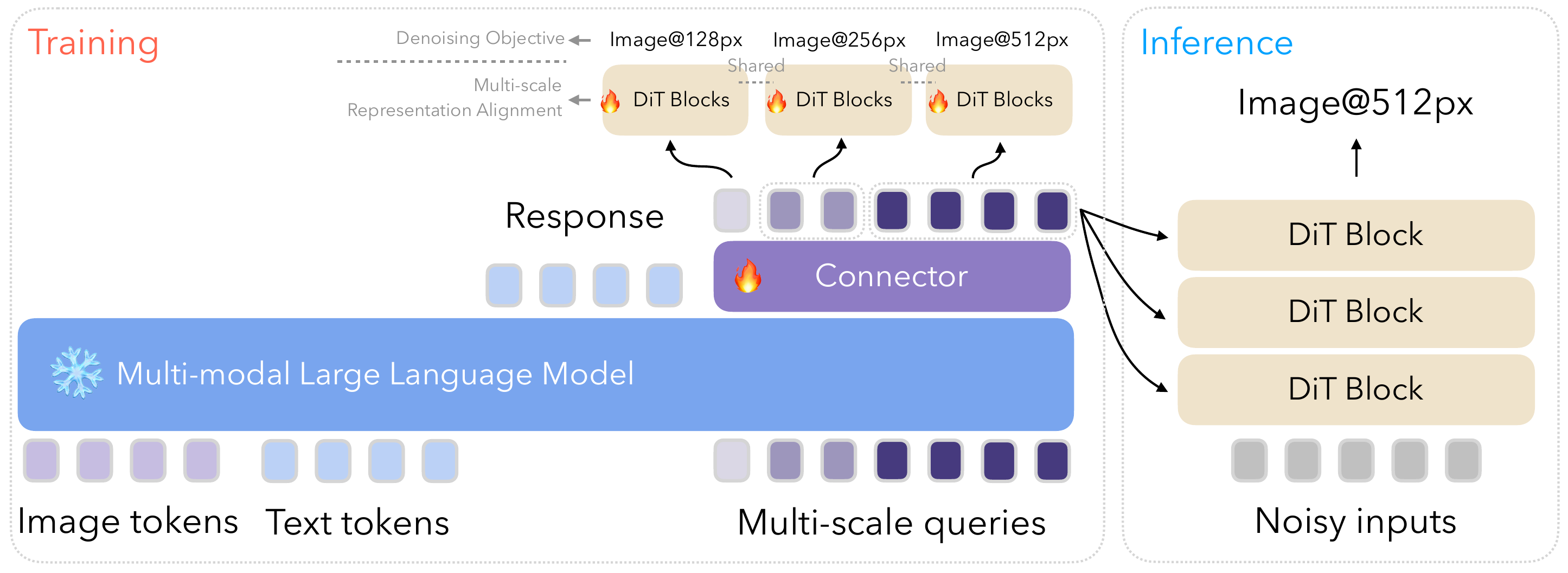}
\caption{\textbf{The framework of \model{}.} Our model fixes the MLLM and fine-tunes the diffusion model through the newly designed multi-scale learnable tokens, multi-scale representation alignment, and connector.}
\label{fig:uae}
\end{figure*}

We propose Multi-Scale Learnable Query Tokens to enhance multi-resolution image understanding and generation within a unified framework. Additionally, we introduce a Multi-Scale Representation Alignment strategy to align intermediate and output representations across different scales. The framework is shown in Fig.~\ref{fig:uae}.

\paragraph{\textbf{Multi-Scale Learnable Tokens Construction}}
Given an input image $x$, we define a set of scales $\mathcal{S} = \{s_1, s_2, \dots, s_K\}$, where each $s_k$ corresponds to a spatial resolution, e.g., $s_k \in \{4\times4, 8\times8, 16\times16\}$. Each scale $s_k$ is associated with a dedicated set of learnable query tokens $Q_{s_k} \in \mathbb{R}^{N_{s_k} \times d}$, where $N_{s_k}$ is the number of tokens for scale $s_k$, and $d$ is the hidden dimension size. Formally, we initialize the multi-scale query tokens as:
\begin{equation}
Q = \{ Q_{s_1}, Q_{s_2}, \dots, Q_{s_K} \}, \quad Q_{s_k} = \text{Learnable Parameters}.
\end{equation}

Each $Q_{s_k}$ is designed to capture information at different granularity levels: (1) Low resolution ($4\times4$), which captures global layout and color distribution. (2) Medium resolution ($8\times8$), which models major objects and mid-level structures. (3) High resolution ($16\times16$), which encodes fine textures and detailed patterns.

\paragraph{\textbf{Multi-Scale Learnable Tokens Fusion and Processing}}
To preserve scale-specific semantics, we introduce explicit scale boundary markers. For each scale $s_k$, we prepend and append special tokens:
\vspace{2mm}
\begin{equation}
\text{Input}_{s_k} = [\text{START}_{s_k}, Q_{s_k}, \text{END}_{s_k}],
\end{equation}
where $\text{START}_{s_k}$ and $\text{END}_{s_k}$ are learnable tokens indicating the boundaries of $s_k$. Each query token also receives a dedicated positional encoding. Let $\mathbf{P}_{s_k} \in \mathbb{R}^{N_{s_k} \times d}$ denote the positional grid encoding for scale $s_k$, constructed according to the spatial grid associated with that resolution. The complete multi-scale token sequence fed into the transformer encoder is thus:
\begin{equation}
\mathbf{Z}_\text{input} = \text{Concat}\left( \text{Input}_{s_1}, \text{Input}_{s_2}, \dots, \text{Input}_{s_K} \right) + \text{Concat}\left( \mathbf{P}_{s_1}, \mathbf{P}_{s_2}, \dots, \mathbf{P}_{s_K} \right).
\end{equation}

The transformer encoder $f_\theta(\cdot)$ processes $\mathbf{Z}_\text{input}$ to produce the hidden representations:
\begin{equation}
\mathbf{H} = f_\theta(\mathbf{Z}_\text{input}).
\end{equation}

\paragraph{\textbf{Multi-Scale Representation Alignment}}
To unify the feature representations used for both image understanding and generation, we introduce a simple yet effective Multi-Scale Representation Alignment strategy (\textit{i.e.,} scale wised consistency loss). Specifically, we align the intermediate hidden states from the DiT backbone with the final semantic representations by minimizing the mean squared error between them.
The alignment loss encourages consistency between  hierarchical representations and final outputs through native-resolution optimization which directly enhances the high-res reconstruction quality (>2dB PSNR) and boosts GenEval by 1.5\%.

\subsection{Native Multimodal Auto-regressive Model (The AR part of \model{})}
\label{subsection:ar}

\begin{figure*}[ht]
\centering
\includegraphics[width=\textwidth]{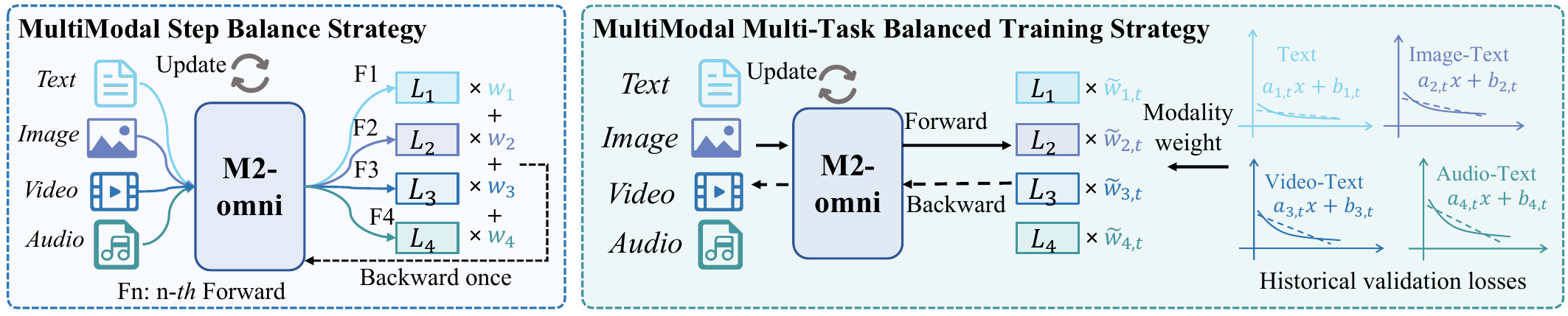}
\caption{\textbf{The AR part of \model{}.} Our model reuses the M2-omni MLLM as a frozen token prediction module, retaining only its text and image branches. The pretraining procedure and dataset of the AR model are consistent with our previous work, please refer to \cite{guo2025m2} for details.}
\label{fig:ae}
\end{figure*}

\paragraph{\textbf{Vision Encoder}}
We utilize a NaViT~\cite{navit_2024} as the vision encoder, capable of processing images of arbitrary resolution. To reduce the length of visual tokens, we concatenate adjacent $2\times2$ tokens into a single token and use an MLP to reduce the dimension to the original dimension, thereby downsampling the visual representation.

\paragraph{\textbf{M2-omni LLM}}
As shown in Fig.~\ref{fig:ae}, the M2-omni LLM integrates the multimodal information and outputs the embedding for multimodal understanding. Our M2-omni LLM is initialized with pre-trained weights from the Llama3~\cite{llama_2023, llama3_2024} series, specifically Llama3.1-8B or Llama3.3-70B. To facilitate unified positional encoding across textual, image, video, and audio modalities, and to enable the model to generalize to longer sequences during inference, we substitute the original 1D-RoPE~\cite{su2024roformer} in Llama with M-RoPE~\cite{qwen2-vl_2024}.

\newpage

\section{Training Data}
\label{sec:data}
The training dataset consists of two parts: basic image-text pairs and image generation data. The basic image-text pairs are primarily sourced from public datasets (\textit{e.g.}, Laion~\cite{laion}, COYO~\cite{byeon2022coyo}) as well as additional images and corresponding captions collected and filtered from the internet (\textit{e.g.}, Midjourney, Google Search). The image generation data mainly comes from commonly used training datasets for various downstream image style transfer tasks or image editing tasks.

\subsection{Basic Image-Text Pairs}

\begin{figure*}[h]
\centering
\includegraphics[width=\textwidth]{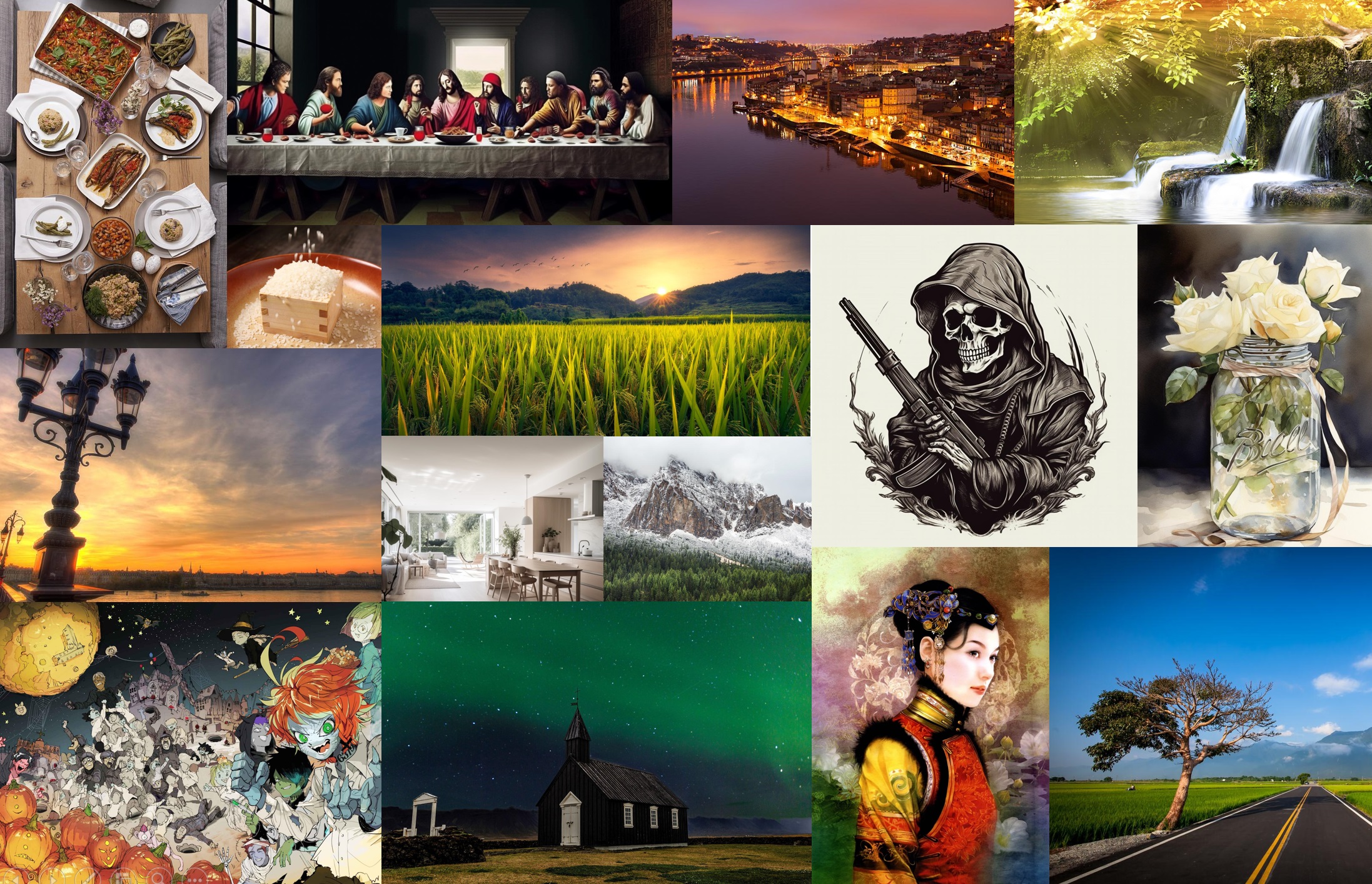}
\caption{\centering Basic image-text training pairs of \model{}.}
\label{fig:dataset}
\end{figure*}

We use LAION-5B (1,555,342,102 samples), Zero~\cite{guo2025m2} (151,766,788 samples), and COYO (61,915,700 samples) as the basic pretrain image-text datasets. Additionally, we collect high-quality image-caption pairs from Wukong~\cite{gu2022wukong}, Midjourney, and web search engines, commonly used for diffusion model training. We apply aspect ratio ($\leq 2.5$), watermark detection ($\leq 0.5$), and CLIP alignment ($\geq 0.45$) thresholds for filtering. After preprocessing, the final sample counts are 34,826,982 (Wukong), 5,421,512 (Midjourney), and 440,951,902 (others). Fig.~\ref{fig:dataset} shows sample visualizations. Finally, our tarining dataset also includes a small amount of aesthetic evaluation data collected from AVA~\cite{6247954} (255,000 samples), TAD66K~\cite{herethinking} (66,000 samples), AesMMIT~\cite{AesExpert} (21,904 samples), and APDD~\cite{jin2024apddv2aestheticspaintingsdrawings} (10,023 samples). Although limited in amount, these data help the model learn human-defined aesthetic standards, improving image generation quality and enhancing the model's ability to assess visual aesthetics.

\subsection{Image Generation Datasets}
\begin{figure*}[h]
\centering
\includegraphics[width=\textwidth]{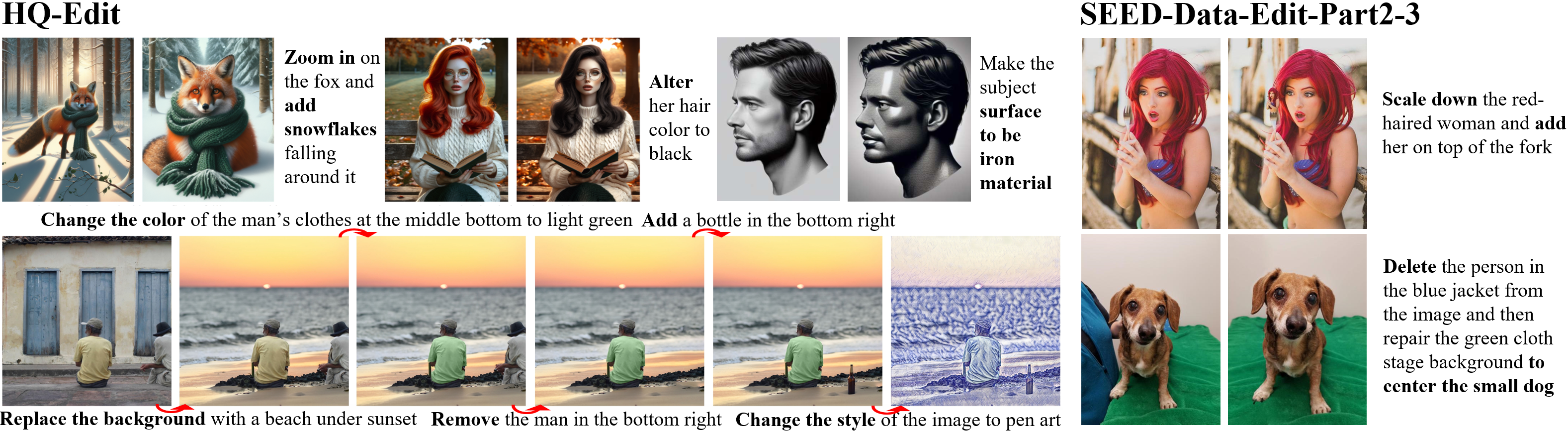}
\vspace{-8mm}
\caption{\centering \model{} image editing data examples in the training set.}
\label{fig:editdataset1}
\vspace{-1mm}
\end{figure*}

This part of training data includes InstructPix2Pix-clip-filtered~\cite{brooks2023instructpix2pix}, where each pair of edited images is generated 100 times, and the best examples are chosem based on CLIP metrics (Sec.3.1.2 in InstructPix2Pix), SEED-Data-Edit-part2/3~\cite{ge2024seededit}, excluding part1 due to the poor visual quality, Ultra-edit~\cite{zhao2024ultraedit}, SynCD~\cite{kumari2025generating}, Subjects200k~\cite{tan2024ominicontrol}, HQ-edit~\cite{hui2024hq}, and MagicBrush~\cite{zhang2023magicbrush}. It consists of 5,008,795 samples. Tab.~\ref{tab:multi_editing} reports the number of sequences with two or more consecutive edits. Representative examples are shown in Fig.~\ref{fig:editdataset1}.

\begin{table*}[h]
\setlength\tabcolsep{2pt}
\def\w{20pt} 
\caption{%
\centering
    Statistics of multi-round editing data. We report the number of samples with 2, 3, 4, and 5 or more consecutive edits.
}
\vspace{-8pt}
\centering\footnotesize
\scalebox{1.2}{
\begin{tabular}{l@{\extracolsep{10pt}}c@{\extracolsep{10pt}}c@{\extracolsep{10pt}}c@{\extracolsep{10pt}}c}
\shline
\textbf{Dataset} & \textbf{2-step Edit} & \textbf{3-step Edit} & \textbf{4-step Edit} & \textbf{5-step Edit and above} \\
\shline
MagicBrush~\cite{zhang2023magicbrush} & 1,151 & 1,572 & - & - \\
SynCD~\cite{kumari2025generating} & 25,438 & - & - & - \\
SEED-Data-Edit-part3~\cite{ge2024seededit} & 472 & 7,453 & 8,783 & 4,669 \\
\shline
\end{tabular}
}
\label{tab:multi_editing}
\end{table*}

\begin{figure*}[h]
\centering
\includegraphics[width=\textwidth]{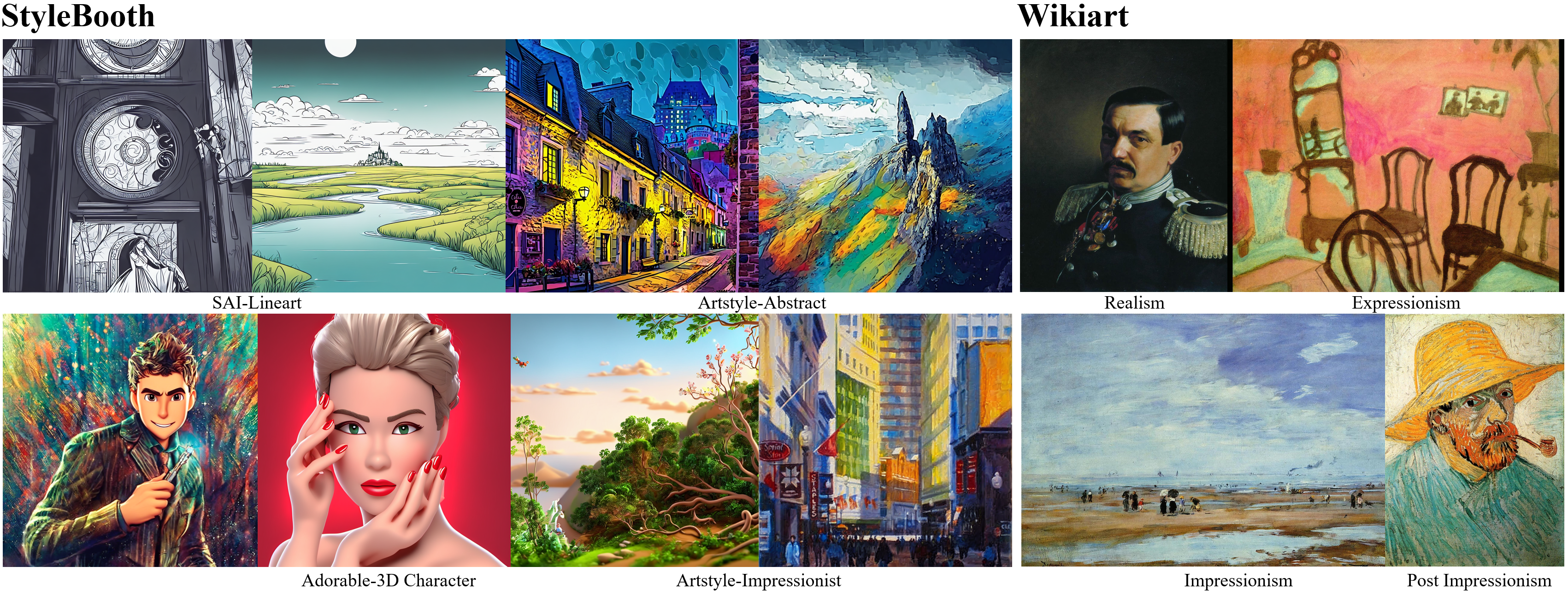}
\vspace{-8mm}
\caption{\centering \model{} image style transfer data examples in the training set.}
\label{fig:editdataset2}
\end{figure*}

In addition, our training data includes publicly available datasets commonly used for style transfer tasks, along with synthesized data generated using style prompts. The style data comprises a high-quality subset of WikiArt, covering 27 painting styles such as Impressionism, Realism, and Expressionism, and the StyleBooth~\cite{han2024stylebooth}, featuring 67 styles including cartoon and 3D, each with 717 image pairs. These two datasets contain 81,444 and 80,922 samples, respectively. Representative examples are shown in Fig.~\ref{fig:editdataset2}.

\section{Benchmark Evaluations}

We conduct separate quantitative evaluations of \model{} on multimodal understanding and text-to-image generation using public benchmarks. For multimodal understanding, we compare against traditional models that take images and text as input and output text, as well as against recent models with visual generative capabilities. For multimodal generation, we evaluate text-to-image performance on GenEval~\cite{ghosh2024geneval}.

\subsection{Multimodal Understanding}
\label{sec:text_evals}

To evaluate the effectiveness of our \model{} in image-text understanding, we benchmark it against state-of-the-art MLLMs on 7 different 
multimodal benchmarks, including complex VQA (MMB~\cite{liu2025mmbench}, MMS~\cite{chen2024we}, MMMU~\cite{yue2023mmmu}, AI2D~\cite{kembhavi2016diagram}, and MM-Vet~\cite{yu2024mm}), multimodal reasoning (MathV~\cite{lu2024mathvista}), and hallucination evaluation (Hall~\cite{Guan_2024_hallusionbench}). Tab.~\ref{tab:exp_it_oc} shows the overall results. Our model achieves top-tier performance on most benchmarks, surpassing closed-source models like GPT-4o and Gemini-1.5-Pro. Furthermore, our model exhibits competitive performance among models of similar size, showcasing its robust capabilities in image-text understanding tasks.

\begin{table}[ht]
    \centering
    \setlength{\tabcolsep}{3pt}
    \renewcommand{\arraystretch}{1.2}
    \scriptsize
    \caption{\textbf{Quantitative results on parts of OpenCompass~\cite{2023opencompass} multimodal leaderboard.}
    $^{\ddag}$ denotes closed-source models. Hall denotes HallusionBench. ``Und.'' and ``Gen.'' denote ``understanding'' and ``generation'', respectively. 
    }
    \label{sota_result_understanding}
    \scalebox{1.2}{
    \begin{tabular}{ll|c|ccccccc}
        \toprule
        \textbf{Type} & \textbf{Model} & \textbf{Avg.} & \textbf{MMB$ \uparrow$} & \textbf{MMS$ \uparrow$} & \textbf{MMMU$ \uparrow$} & \textbf{MathV$ \uparrow$}  & \textbf{Hall$ \uparrow$}  & \textbf{AI2D$ \uparrow$}  & \textbf{MM-Vet$ \uparrow$} \\
        \midrule
        \textit{Und. Only} & 
        LLaVA-72B~\cite{xie2024show} & 68.0  & 84.5  & 65.8  & 56.6  & 68.4  & 47.9  & 86.2   & 60.6 \\
        & Qwen2-VL-72B~\cite{bai2023qwen} & 74.8  & 85.9  & 68.6  & 64.3  & 69.7  & 58.7  & 88.3   & 73.9 \\
        & Qwen2.5-VL-7B~\cite{bai2025qwen25vltechnicalreport} & 76.2  & 87.8  & 71.1  & 67.9  & 70.8  & 58.8  & 88.2    & 76.7 \\
        & Emu$3$-Chat~\cite{wang2024emu3} & - & 58.5 & - & 31.6 & - &  - & - & 37.2 \\     
        & InternVL2.5-8B~\cite{chen2024expanding} & 70.3  & 82  & 65.2  & 54.8  & 67.9  & 51.7  & 84.5   & 68.1 \\  
        & InternVL2.5-38B~\cite{chen2024expanding}  & 73.5  & 85.4  & 68.5  & 64.6  & 72.4  & 57.9  & 87.6   & 67.2 \\   
        & InternVL2.5-78B~\cite{chen2024expanding}  & 75.2 & 87.5  & 69.5 & 70 & 71.4 & 57.4 & 89.1 & 71.8 \\   
        & DeepSeek-VL2~\cite{wu2024deepseekvl2}  & 66.4  & 81.2  & 61.0  & 50.7  & 59.4  & 51.5  & 84.5  & 60.0 \\   
        & GPT-4o-20241120$^{\ddag}$~\cite{openai2024gpt4ocard} & 72.0   & 84.3  & 65.1  & 70.7  & 59.9  & 56.2  & 84.9  & 74.5 \\   
        & Step-1o$^{\ddag}$~\cite{step1o} & 77.7  & 87.3  & 69.3  & 69.9 & 74.7  & 55.8 & 89.1 & 82.8 \\   
        \midrule
        \textit{Und. and Gen.} 
        & DreamLLM~\cite{dong2023dreamllm} & - & - & - & - & - &  - & - & 36.6 \\
        & MetaMorph~\cite{tong2024metamorph} & - & 75.2 & - & - & - &  - & - & - \\
        & Show-o-256~\cite{xie2024show} & - & - & - & 25.1 & - &  - & - & - \\
        & Show-o-512~\cite{xie2024show} & - & - & - & 26.7 & - &  - & - & - \\
        & TokenFlow-XL~\cite{qu2024tokenflow} & - & 68.9 & - & 38.7 & - &  - & - & 40.7 \\
        & Chameleon~\cite{team2024chameleon} & - & - & - & 22.4 & - &  - & - & 8.3 \\
        & Janus~\cite{deepseek_janus} & - & 69.4 & - & 30.5 & - &  - & - & 34.3 \\
        & Janus-Pro-7B~\cite{chen2025janus} & - & 79.2 & - & 41.0 & - &  - & - & 50.0 \\
        \cdashline{2-10}
        & \textbf{Ours (\modellite{})} & 69.7  & 80.7  & 60.5  & 51.2  & 68.3  & 51.8 & 84.5  & 72.3 \\
        \bottomrule
    \end{tabular}
    }
    \label{tab:exp_it_oc}
\end{table}

\subsection{Text-to-Image Generation}
\label{sec:image2text_evals}

We report visual generation performance on GenEval~\cite{ghosh2024geneval}. As shown in Tab.~\ref{tab:geneval}, our \model{} obtains 0.62 overall accuracy on GenEval, which outperforms all the other unified or generation-only methods, \textit{e.g.,} MetaQueries~\cite{pan2025transfer} (0.61), DALL-E 3~\cite{dalle3} (0.67), and SDXL~\cite{podell2023sdxl} (0.55). 
These results demonstrate that our model matches the performance of state-of-the-art diffusion models in generating single-subject images.

\begin{table}[ht]
    \centering
    \setlength{\tabcolsep}{2.0pt}
    \renewcommand{\arraystretch}{1.2}
    \scriptsize
    \caption{\textbf{Evaluation of text-to-image generation ability on GenEval benchmark~\cite{ghosh2024geneval}}. ``Und.'' and ``Gen.'' denote ``understanding'' and ``generation'', respectively.
    }
    \vspace{-2mm}
    \scalebox{1.2}{
    \begin{tabular}{llcccccc|c}
        \toprule
        \textbf{Type} & \textbf{Method}  & \textbf{Single Obj.$\uparrow$} & \textbf{Two Obj.$\uparrow$} & \textbf{Counting$\uparrow$} & \textbf{Colors$\uparrow$} & \textbf{Position$\uparrow$} & \textbf{Color Attri.$\uparrow$} & \textbf{Overall$\uparrow$} \\
        \midrule
        \multirow{8}{*}{\textit{Gen. Only}} 
        & LlamaGen~\cite{sun2024autoregressive}  & 0.71 & 0.34 & 0.21 & 0.58 & 0.07 & 0.04 & 0.32 \\
        & LDM~\cite{rombach2022high} & 0.92 & 0.29 & 0.23 & 0.70 & 0.02 & 0.05 & 0.37 \\
        & SDv$1.5$~\cite{rombach2022high} &  0.97 & 0.38 & 0.35 & 0.76 & 0.04 & 0.06 & 0.43 \\
        & PixArt-$\alpha$~\cite{chen2023pixart} &  0.98 & 0.50 & 0.44 & 0.80 & 0.08 & 0.07 & 0.48 \\
        & SDv$2.1$~\cite{rombach2022high} &  0.98 & 0.51 & 0.44 & 0.85 & 0.07 & 0.17 & 0.50 \\
        & DALL-E $2$~\cite{ramesh2022hierarchical}  & 0.94 & 0.66 & 0.49 & 0.77 & 0.10 & 0.19 & 0.52 \\
        & Emu$3$-Gen ~\cite{wang2024emu3}  & 0.98 & 0.71 & 0.34 & 0.81 & 0.17 & 0.21 & 0.54 \\
        & SDXL~\cite{podell2023sdxl} &  0.98 & 0.74 & 0.39 & 0.85 & 0.15 & 0.23 & 0.55 \\
        & DALL-E $3$~\cite{dalle3}  & 0.96 & 0.87 & 0.47 & 0.83 & 0.43 & 0.45 & 0.67 \\
        & SD3-Medium~\cite{esser2024scalingrectifiedflowtransformers} & 0.99 & 0.94 & 0.72 & 0.89 & 0.33 & 0.60 & 0.74 \\
        \midrule
        \multirow{5}{*}{\textit{Und. and Gen.}}
        & Chameleon~\cite{team2024chameleon} &  - & - & - & - & - & - & 0.39 \\
        & LWM~\cite{liu2024world} &  0.93 & 0.41 & 0.46 & 0.79 & 0.09 & 0.15 & 0.47 \\
        & SEED-X~\cite{ge2024seed}  & 0.97 & 0.58 & 0.26 & 0.80 & 0.19 & 0.14 & 0.49 \\
        & Show-o~\cite{xie2024show} &  0.95 & 0.52 & 0.49 & 0.82 & 0.11 & 0.28 & 0.53 \\
        & TokenFlow-XL~\cite{liu2024world} &  0.95 & 0.60 & 0.41 & 0.81 & 0.16 & 0.24 & 0.55 \\
        & Janus~\cite{wu2024janus} & 0.97 & 0.68 & 0.30 & 0.84 & 0.46 & 0.42 & 0.61\\
        & Janus-Pro-1B~\cite{chen2025janus} &  0.98 & 0.82 & 0.51 & 0.89 & 0.65 & 0.56 & 0.73 \\
        & MetaQueries~\cite{pan2025transfer} &  - & - & - & - & - & - & 0.61 \\
        \cdashline{2-9}
        & \textbf{Ours (\modellite{})} &  0.99 & 0.76 & 0.53 & 0.87 & 0.26 & 0.30 & 0.62 \\

        \bottomrule
    \end{tabular}
}
    \label{tab:geneval}
\vspace{-2mm}
\end{table}

\subsection{Instruction Based Image Editing}
\label{sec:editing}

\begin{figure*}[ht]
\centering
\vspace{-2mm}
\includegraphics[width=0.97\textwidth]{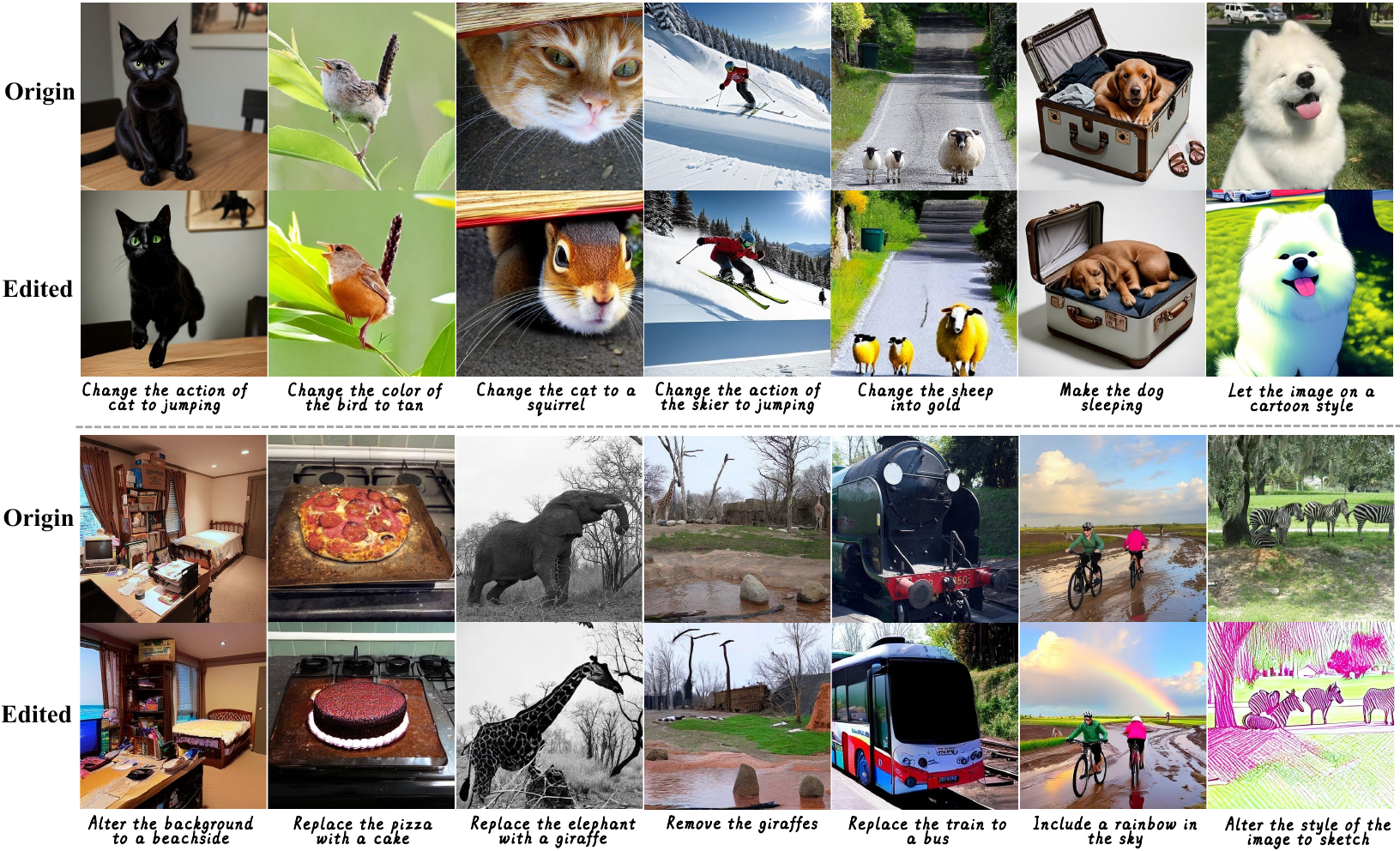}
\vspace{-2mm}
\caption{\centering Instruction based image editing results outputted by \model{}.}
\label{fig:editing}
\vspace{-30mm}
\end{figure*}

As shown in Fig.~\ref{fig:editing}, we conducted a qualitative analysis on a wider range of interactive image editing tasks, including style transfer and object addition, deletion, and modification. 

\newpage
\section{Contributors}
\label{sec:contri}

\large{Authors are listed \textbf{alphabetically by the first name}.}

\large{
\begin{multicols}{3}
\raggedcolumns

Biao Gong\\
Cheng Zou\\
Dandan Zheng\\
Hu Yu\\
Jingdong Chen\\
Jianxin Sun\\
Junbo Zhao\\
Jun Zhou\\
Kaixiang Ji\\
Lixiang Ru\\
Libin Wang\\
Qingpei Guo\\
Rui Liu\\
Weilong Chai\\
Xinyu Xiao\\
Ziyuan Huang\\

\end{multicols}}

\clearpage

\bibliographystyle{assets/plainnat}
\bibliography{main}

\end{document}